\documentclass[journal]{IEEEtran}
\usepackage{cite}
\ifCLASSINFOpdf
  \usepackage[pdftex]{graphicx}
  \DeclareGraphicsExtensions{.pdf,.jpeg,.png,.eps}
\else
  \usepackage[dvips]{graphicx}
  \DeclareGraphicsExtensions{.pdf}
\fi
\usepackage[caption=false, font=footnotesize]{subfig}
\usepackage{amsmath}
\usepackage{amssymb}
\usepackage{booktabs}
\usepackage{mdwtab}

\usepackage{float}
\usepackage{placeins}
\usepackage{ragged2e}
\usepackage{url}

\usepackage{algorithm, algorithmic}

\usepackage{pgfmath}
\usepackage{color}

\usepackage[unicode=true,
 bookmarks=true,bookmarksnumbered=true,bookmarksopen=true,bookmarksopenlevel=1,
 breaklinks=false,pdfborder={0 0 0},pdfborderstyle={},backref=false,colorlinks=false]
 {hyperref}

\makeatletter

\begin{document}

\title{An Efficient Deep Reinforcement Learning Model for Urban Traffic Control}

\author{Yilun~Lin,
        Xingyuan Dai,
        Li~Li,
        and~Fei-Yue Wang
\thanks{This work was supported in part by National Natural Science Foundation of China (Grant No. 61533019, 71232006),
the Beijing Municipal Science and Technology Commission Program under Grant D171100000317002,
the Beijing Municipal Commission of Transport Program under Grant ZC179074Z. (Corresponding author is \textit{Li Li})}
\thanks{Y. Lin, X. Dai and F.-Y. Wang are with the State Key Laboratory for Management and Control of Complex Systems, Institute of Automation, Chinese Academy of Sciences, Beijing 100080, China(Email: \{linyilun2014, daixingyuan2015, feiyue.wang\}@ia.ac.cn) .}
\thanks{Y. Lin, X. Dai are also with University of Chinese Academy of Sciences, Beijing 100049, China.}
\thanks{L. Li is with Department of Automation, TNList, Tsinghua University, Beijing 100084, China (Tel: +86(10)62782071, Email: li-li@tsinghua.edu.cn).}
\thanks{All authors are also affiliated with Qingdao Academy of Intelligent Industries, Qingdao, Shandong, 266109, China.}
}

\maketitle
\begin{abstract}
Urban Traffic Control (UTC) plays an essential role in Intelligent Transportation System (ITS) but remains difficult. Since model-based UTC methods may not accurately describe the complex nature of traffic dynamics in all situations, model-free data-driven UTC methods, especially reinforcement learning (RL) based UTC methods, received increasing interests in the last decade. However, existing DL approaches did not propose an efficient algorithm to solve the complicated multiple intersections control problems whose state-action spaces are vast. To solve this problem, we propose a Deep Reinforcement Learning (DRL) algorithm that combines several tricks to master an appropriate control strategy within an acceptable time. This new algorithm relaxes the fixed traffic demand pattern assumption and reduces human invention in parameter tuning. Simulation experiments have shown that our method outperforms traditional rule-based approaches and has the potential to handle more complex traffic problems in the real world.
\end{abstract}


\IEEEpeerreviewmaketitle{}

\section{Introduction}
\IEEEPARstart{U}{rban} Traffic Control (UTC) systems aim to better schedule vehicles' movements, exploit the capacity of existing road networks and mitigate traffic congestion in urban areas without significant cost. However, it remains challenging to design an appropriate UTC system, since it is hard to accurately describe the complex nature of urban traffic networks to find proper signal timing plans.

Early UTC systems were mainly built on some simplified traffic flow models and under the assumption of relatively fixed traffic demand patterns within in a short period \cite{Hunt-1981}. However, the success of such approaches relies on the tedious adjustment of experienced transportation engineers. Moreover, the correlations between intersections often vary noticeably from time to time and thus make the pre-defined signal timing plan not optimal.

To solve this problem, model-free data-driven UTC methods, especially reinforcement learning (RL) based UTC methods, received increasing interests in the last decade, along with the fast development of artificial intelligence theory and intelligent control techniques.
Instead of optimizing signal timing plan according to simplified traffic flow models, these approaches aim to self-learn the optimal timing policy by analyzing thousands of samples between the change of traffic states and control actions.
The invention of human experts in parameter tuning could be replaced by online learning, too.

Among various model-free data-driven UTC methods, Reinforcement Learning (RL) based traffic control receives increasing attention \cite{Abdulhai-2003a}, since RL has been successfully used in many applications other than traffic control. In general, Reinforcement Learning allows the system to learn how to choose its behaviors based on feedback from the environment. For traffic control problems, RL based approaches usually take the traffic flow states around the intersections as the observable states, the change of signal timing plans as actions, and the traffic control performance as feedback. After transformation, the traffic control problem will be treated as a standard RL problem and solved by using some standard RL algorithms.

Initial RL based approaches considered the signal timing for isolated intersections \cite{El-Tantawy-2014}.
Most of them consist of a classical algorithm like Q-Learning \cite{Watkins-1992} and SARSA \cite{Rummery-1994} to control the timing of a single intersection \cite{Abdulhai-2003, Thorpe-1997, Richter-2007, Shoufeng-2008a, Salkham-2008}.
Conventional RL based approaches used tables to record and describe the relationship between the states and actions.
As a result, it is difficult to use them for UTC problem with multiple intersections, since the dimension of state-action spaces is too vast to learn.

One solution to this problem is to apply divide-and-conquer policy: divide the studied road network region into small grids containing a few intersections and then solve the traffic control problem for each grid respectively, in the lower-level.
In the upper-level, each grid is treated as an agent and is allowed to cooperate to seek a globally optimal solution \cite{Wiering-2000, Camponogara-2003, deOliveira-2006, BoChen-2010,Balaji-2010a,Arel-2010,Caselli-2015}.
However, multi-agent approaches ease the difficulties of optimization while introducing other problems.
For example, it is hard to obtain a real global optimal global control, since each agent usually can only receive limited information \cite{Li-2016b}.

Another solution is to directly attack UTC problems with multiple intersections by using some advanced algorithms to overcome the curse of dimension.
For example, Deep Learning (DL) \cite{LeCun-2015}, as one of the most recent and successful breakthroughs in AI research, has been introduced and combined with RL methods.
The benefit of DL lies in its capability to quickly learn and capture the relationship between the states and actions by using a data structure (deep neural networks) that is more efficient than tables.
The integration of DL and RL, widely known as Deep Reinforcement Learning (DRL), has already shown its potential by successfully solving video games \cite{Mnih-2015}, 3D locomotion \cite{Heess-2017}, Go game \cite{Silver-2017} and many other problems.

One of the earliest attempts to solve traffic control problem via DRL methods proposed by Li et al. \cite{Li-2016a} used the Deep Q-Network \cite{Mnih-2015} to control a single intersection.
In the follows, researchers had extended such method by applying it to different scenarios,  such as traffic light coordination \cite{vanderPol-2016}. Such methods have also been improved by proposing new traffic state encoding methods \cite{Genders-2016}, or using different models such as Deep Deterministic Policy Gradient \cite{Casas-2017a}.

However, existing DRL based UTC models do not always work well in scenarios with multiple intersections because of the following shortcomings. First, some deep neural networks (e.g., the Deep Q-Network applied in \cite{Li-2016a}) used for model the relationship between the states and the actions do not fit for large-scale UTC problems that contain multiple intersections. Second, some reward functions recommended for RL do not appropriately characterize the desired state of traffic systems when the correlations between intersections become highly interlaced. Third, some algorithms designed for the training of DRL based UTC models cannot keep a proper balance between solution space exploration and optimal solution seeking. These algorithms are too slow to reach a satisfactory solution for large-scale UTC problems.

To solve these problems, we propose an efficient DRL model dedicated to large-scale UTC problems.
First, it uses Residual Networks (ResNet) \cite{He-2016} as the deep neural network model to learn the relationship between the states and the actions.
Second, we test different reward functions and design a hybrid reward, in which the
throughput of the traffic network, along with the balance of queueing length
around intersections is chosen as the performance indexes.
Third, it applies a new policy update algorithm, called clipped Proximal Policy Optimization (PPO) algorithm.
Moreover, we allow this new model to work with the relaxed traffic demand pattern assumption and
the human invention in parameter tuning is significantly reduced.

Tests show that this new model could be optimized within an acceptable time for a traffic grid. Compared with previous DRL models which take thousands of episodes to converge, our method takes only less than $50$ episodes to converge for a more complex environment. The entire training stage took only several hours on a workstation with two GPUs. Simulation results show that this deep learning powered UTCS can increase the average capacity of traffic system by $10.91\%$ while reducing the average waiting time by $15.57\%$ compared with the fixed-time controller.

\begin{figure}[h]
\begin{centering}
\includegraphics[width=1\columnwidth]{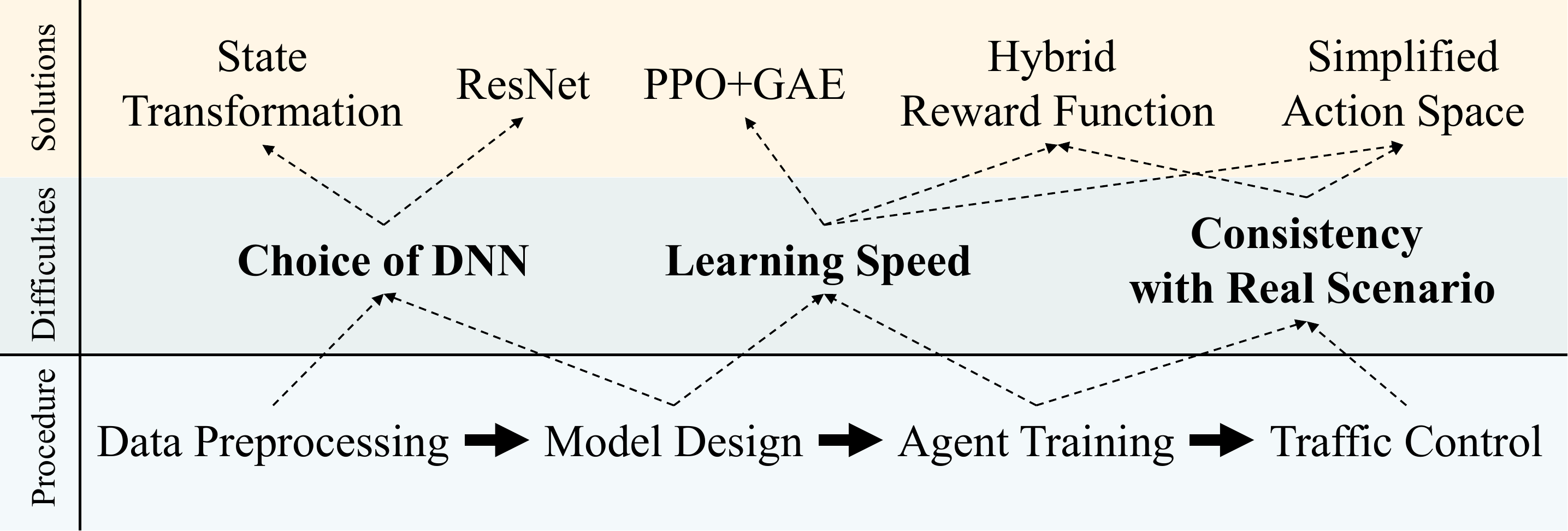}
\par\end{centering}
\caption{The major problems addressed in this paper and our contributions.\label{fig:contributions}}
\end{figure}

Fig. \ref{fig:contributions} shows the techniques that we used to handle the interlaced difficulties. To better explain our findings, we organize the paper in the following way. First, we will briefly introduce the background of reinforcement learning in \emph{Section \ref{sec:rl-background}} for further discussion. Then, we will present how to consider a UTC problem from the viewpoint of DRL in \emph{Section \ref{sec:method}}. Simulation results will be demonstrated in \emph{Section \ref{sec:results}}. Finally, we conclude our contributions  and discuss some future applications in \emph{Section \ref{sec:conclusions}}.

\section{Reinforcement Learning Background\label{sec:rl-background}}

To better present our findings, it is necessary to briefly review the basic idea of reinforcement learning in this  section and list the terms/symbols that will be used in the follows; see Table \ref{tab:notations}.

\begin{table}[htbp]
  \centering
  \caption{Summary of Notations}  \label{tab:notations}%
    \begin{tabular}{c|p{0.7\columnwidth}}
    Symbol & Meaning \\
    \hline
    $s_t$ & State of the environment at time step $t$ \\
    $a_t$ & Action taken by the agent at time step $t$ \\
    $r_t$ & Immediate return given by the environment for $a_t$ \\
    $R_t$ & The overall return given by the environment at time step $t$ \\
    $\pi$ & The policy \\
    $V(s_t)$ & The value of $s_t$, which is the overall return on an infinite time horizon since time step $t$ \\
    $Q(s_t, a_t)$ & The Q-value of $s_t$ by taking action $a_t$ \\
    $A_t$ & The abbreviations of the advantage $A(s_t, a_t)$ \\
    $\theta$ & The parameters of policy/critic model \\
    \end{tabular}%
\end{table}%

In RL problems, we assume that an agent interacts with an environment $\mathcal{E}$ over a number of discrete time steps to maximize the reward \cite{Sutton-1998}.
An RL problem is often represented by a quintuple
$\left\langle\mathcal{S},\mathcal{A},\mathcal{P}_{a}\left(s,s^{\prime}\right),\mathcal{R}_{a}\left(s,s^{\prime}\right),\mathcal{\gamma}\right\rangle
$, where $\mathcal{S}$ is a set of states, $\mathcal{A}$ is a set of possible
actions. $\mathcal{P}_{a}\left(s,s^{\prime}\right)$ is the probability that action
$a$ will lead to state $s^{\prime}$ from state $s$ in time step $t$, and
$\mathcal{R}_{a}\left(s,s^{\prime}\right)$ is the corresponding expected
immediate reward. $\gamma\in\left[0,1\right]$ is the discount factor, which
represents the difference in importance between future rewards and present
ones.

Our goal is to choose a policy function $\pi$ that will maximize some cumulative function of the random rewards, typically the expected discounted sum over a potentially infinite horizon from each state $s_{t}$:
\begin{equation}
R_{t}=\sum_{k=0}^{\infty}\gamma^{k}r_{t+k}\label{eq:expected-dis-reward}
\end{equation}

The policy function $\pi$ is usually defined as a mapping from the state $s_{t}$ to the action $a_{t}$.
Because we usually do not know the state transition probability function $\mathcal{P}_{a}\left(s,s^{\prime}\right)$ in advance, we learn $\mathcal{P}_{a}\left(s,s^{\prime}\right)$ and meanwhile seek the optimal policy by trial-and-error search.
At each time step $t$, the agent receives a state $s_{t}$ , and selects an action $a_{t}$ according to its policy $\pi$. In return, the agent receives the next state $s_{t+1}$ and receives a reward signal $r_{t}$.
The process continues until the agent reaches a terminal state after which the process restarts.

To find the desired policy function recursively, we introduce the action value function and the value function.
The action value function $Q^{\pi}(s,a)=\mathbb{E}[R_{t}|s_{t}=s,a]$ is the expected return for selecting action $a$ following policy $\pi$ in state $s$.
The optimal action value function $Q^{\ast}(s,a)=\max_{\pi}Q^{\pi}(s,a)$ gives the maximum action value for state $s$ and action $a$ achievable by any policy.

Similarly, the value of state $s$ under policy $\pi$ is defined as $V^{\pi}(s)=\mathbb{E}[R_{t}|s_{t}=s]$, which is the expected return for following policy $\pi$ from state $s$. The optimal value function $V^{*}\left(s\right)=\max_{\pi}V^{\pi}\left(s\right)$ gives the maximum value for state $s$ achievable by any policy.

In this paper, we consider the neural network-based value function and the associated policy-based methods \cite{Schulman-2017}.
Here, we parameterize the policy as $\pi(a|s;\theta)$ and update the parameters $\theta$ to maximize the cumulative return.
By performing the approximate gradient ascent on $\mathbb{E}\left[R_{t}\right]$,
the parameterized policy $\pi(a|s;\theta)$ tends to choose the action $a$ that maximizes future return from state $s$.
One of the earliest algorithms for such methods, called REINFORCE algorithm \cite{Williams-1992} updates the policy parameters $\theta$ in the direction $\nabla_{\theta}\log\pi\left(a_{t}|s_{t};\theta\right)R_{t}$, which is an unbiased estimate of $\nabla_{\theta}\mathbb{E}\left[R_{t}\right]$.

However, the REINFORCE family of algorithms are still time-consuming when the state-action space is large to explore and learn.
Most recent works use a variant of this approach called Advantage Actor-Critic (A2C) architecture \cite{Sutton-1998,Degris-2012}.
In this architecture, we do not strictly follow the direction indicated by the gradient ascent of $\mathbb{E}\left[R_{t}\right]$.
Instead, we consider the policy gradient scaled by \emph{advantage} $A_t\left(a_{t},s_{t}\right)$ instead of cumulative return $R_{t}$.

The advantage $A_t\left(a_{t},s_{t}\right)$ is calculated using the return subtracting a learned \emph{baseline} function $b_{t}\left(s_{t}\right)$. $b_{t}\left(s_{t}\right)$ can be interpreted as excessive profit gained by taking action $a_{t}$ in state $s_{t}$. In such setting, the policy $\pi$ is viewed as the actor and the baseline $b_{t}$ is viewed as the critic.
The resulting gradient is estimated as

\begin{equation}
\nabla\theta=\hat{\mathbb{E}}\left[\nabla_{\theta}\log\pi_{\theta}\left(a_{t}|s_{t}\right)A_{t}\right]\label{eq:policy_gradient}
\end{equation}
where the expectation $\hat{\mathbb{E}}\left[\dots\right]$ indicates the empirical average over a batch of samples.

In this paper, we follow the above idea but use several improved algorithms which yield significantly faster training times and higher data efficiency in many applications. We will present the details of these algorithms later in the \emph{Section \ref{subsec:Learning Algorithm}}.

\section{RL-Based Urban Traffic Control System\label{sec:method}}

In this section, we explain how to build an urban traffic control system using the reinforcement learning method.
Instead of designing an RL model that can be used in a specific situation but cannot be generalized well, we aim to provide an architecture that can handle most cases with little adjustment.

\subsection{State Space\label{subsec:State-Space}}

In this paper, the data obtained from different sensors are formatted into a 2-D $H\times W$ tensor.
More precisely, we format the data collected at the time step $t$ into a triple $\left\langle C, H, W\right\rangle $, where $C$ is the number of channels, $H$ is the height of input tensor, and $W$ the width of input tensor.

\begin{figure}[h]
\begin{centering}
\textsf{\includegraphics[width=0.9\columnwidth]{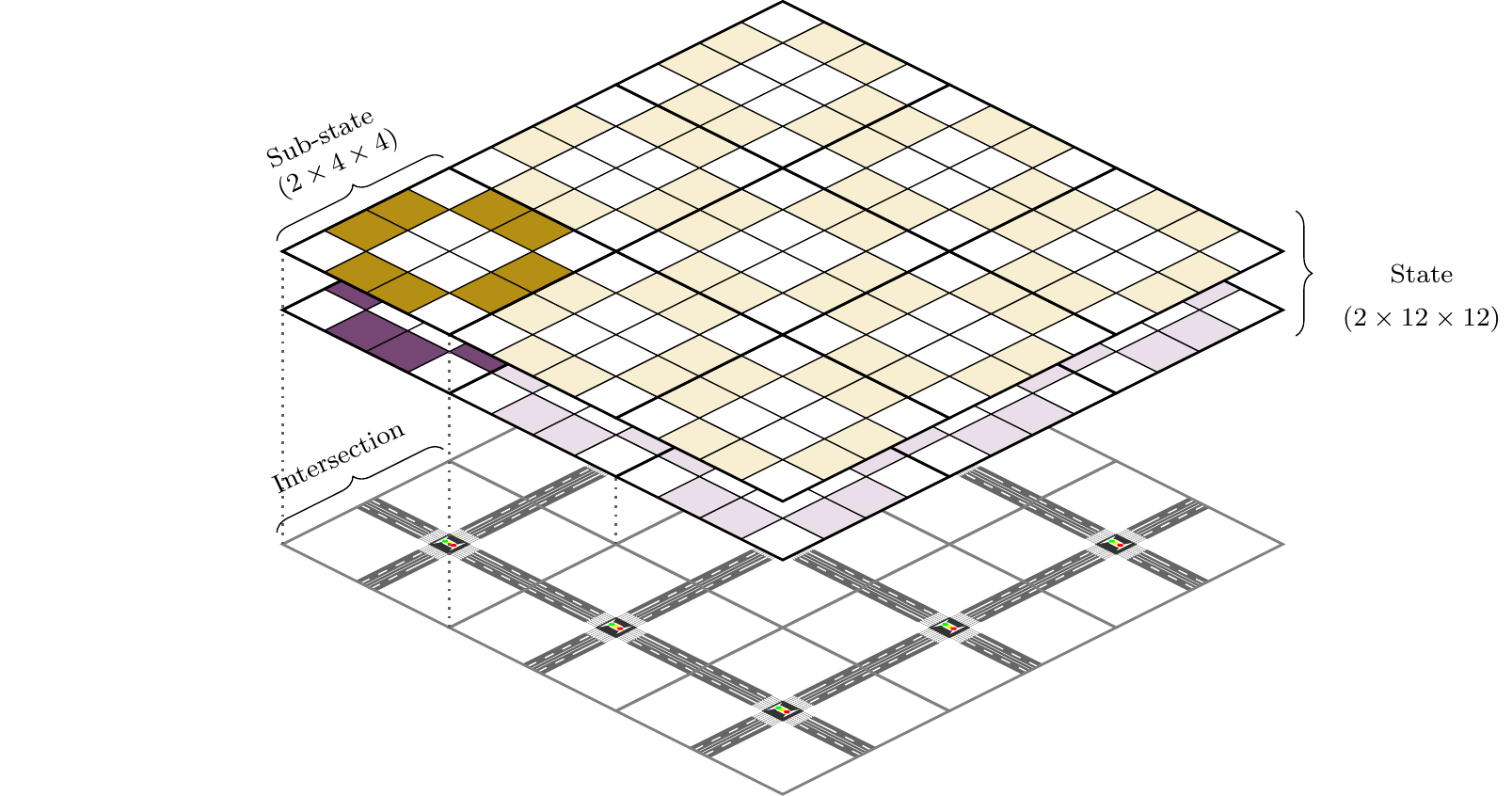}}
\par\end{centering}
\caption{The traffic grid and corresponding formatted tensor.\label{fig:format-of-state}}
\end{figure}

For example, let us consider the road network illustrated in the bottom of Fig. \ref{fig:format-of-state} in the rest of this paper.
In this $3\times3$ grid with $9$ intersections, each intersection has $4$ arms whose length is $500$ meters. Eight sensors are placed on each traffic light to monitor the halting vehicle number and the mean speed. Each sensor is capable of monitoring $150$ meters length at most.

Since two types of information: the halting vehicle number and the mean speed are collected in each intersection, the sub-state can then be formatted into a $2\times4\times4$ tensor as shown in the upper left of Fig. \ref{fig:format-of-state}. The blank cell indicates zero-padding operations.
Therefore, the complete state $s_{t}$ agent received is in a shape of $\left\langle 2,12,12\right\rangle $.

\subsection{Action Space}

The setting of action space is critical to the successful applications of the RL model.
Most previous works use a discrete action space, in which the agent chooses a phase from all possible phases to execute in every time step $t$.
In this paper, we use a similar but simplified action space.

\begin{figure}[h]
\begin{centering}
\includegraphics[width=0.6\columnwidth]{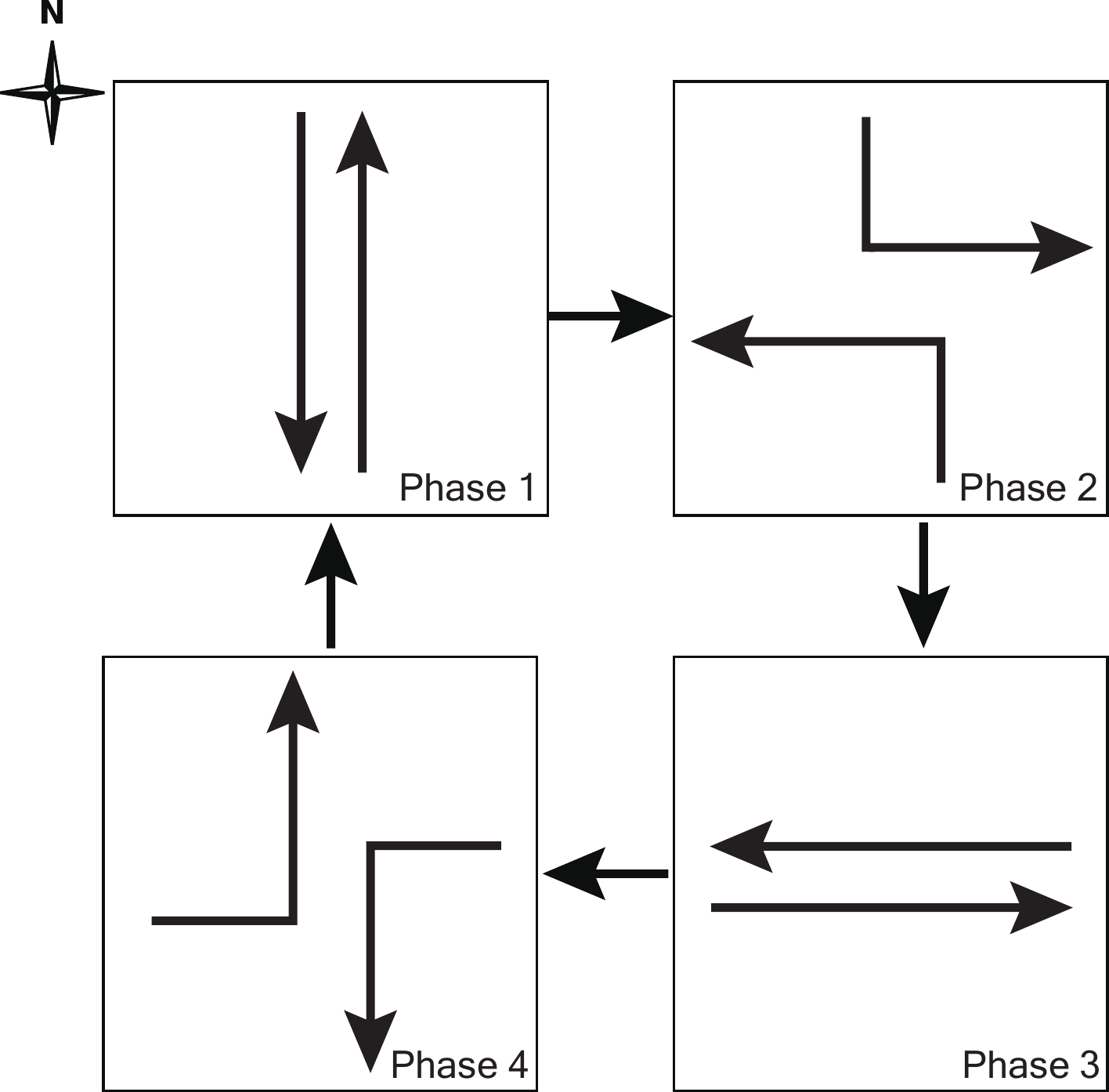}
\par\end{centering}
\caption{$4$ phases of traffic light, \textbf{Straight(NS,SN)}, \textbf{TurnLeft(NE,SW)}, \textbf{Straight(WE,EW)}, \textbf{TurnLeft(WN,ES)} in each intersection. Turn right is always allowed and not shown in this figure. \label{fig:phases}}
\end{figure}

For each intersection, we predefine the possible phases and the order; see Fig. \ref{fig:phases} for a demonstration.
Specially, in this paper, we assume that a yellow light phase lasting for $3$ seconds will be first applied if the traffic light switch from green to red.

For every second, the agent can choose either maintain current phase or switch to next one once the minimal phase duration time (5 seconds in this paper) has passed. Since we are using a centralized way to build the controller, the output of actor is in a shape of $\left\langle N_{\text{TLS}},2\right\rangle $, where 2 indicates the number of the discrete probabilities of choices: either maintaining or switching current phase for each intersection.
Here, $N_{\text{TLS}}=9$ is the number of traffic lights.

\subsection{Reward Function}

Unlike the score in many game scenarios, there is no concise yet perfect indicator of the traffic control performance.
Generally, it is essential to make the reward reflect the nature of the optimal policy.
Meanwhile, it is also vital to avoid the sparse or fluctuate reward signal that is unexpected in a smooth and acceptable learning process.

Various performance indices (e.g., the change in the number of queued vehicles,
the change in cumulative vehicle delay, the change in vehicle throughput or the imbalance between different arms of each intersection) had been used to evaluate the traffic system during the last two decades.

In this paper, we divide the reward signal into two parts.

One part is called as the global reward that can lead the agent to learn optimal strategy to maximize the capacity of the whole road network \cite{Lin-2014}.
More precisely, we choose the net outflow of the road network as the global part of the reward.
The net outflow is calculated by subtracting  the income volume $\|\text{Veh}^{\text{(in)}}_t\|$ from the outcome volume $\|\text{Veh}^{\text{(out)}}_t\|$ within the selected area at each time step $t$:

\begin{equation}
r_{t}^{\text{Global}}=\|\text{Veh}^{\text{(out)}}_t\|-\|\text{Veh}^{\text{(in)}}_t\|\label{eq:global_reward}
\end{equation}

Noted that our experiments are conducted in simulated environments, in which vehicles may teleport (be removed from the network immediately) due to congestions or collisions, we count the outcome volume $\|\text{Veh}^{\text{(out)}}_t\|$ without the teleporting vehicles in the experiments.

The other part is called as the local reward that urges the agent to learn to balance the traffic situation for each intersection.
Though the local reward does not relate with the capacity of the road network directly, it has been proved to be useful for improving the performance of the controller in many works \cite{Ng-1999, VanSeijen-2017}.

The local part of the reward signal also helps to stabilize the agent behavior.
Therefore, we choose the opposite of absolute imbalance of each intersection as some previous work did \cite{Lin-2011a, Tong-2015, Li-2016b}.
It is defined as the absolute negative difference between queue length in north-south/south-north direction and those in east-west/west-east direction, i.e.

\begin{equation}
r_{t}^{\text{TLS}_{i}}=-\left|\max q_{t}^{\textrm{WE}}-\max q_{t}^{\textrm{NS}}\right|\label{eq:local_reward}
\end{equation}

For each intersection $\text{TLS}_{i}$, $q_{t}^{\textrm{WE}}$ is the number of halting vehicle in lanes from west to east or vice versa. Similarly, $q_{t}^{\textrm{NS}}$ is that from north to south or vice versa.

The complete hybrid reward function can then be formed by summing up the global and local parts.
\begin{equation}\label{eq:reward}
  r_{t} = \beta r_{t}^{\text{Global}} + (1-\beta) \frac{1}{N_{\text{TLS}}}\sum_{i}^{N_{\text{TLS}}}{r_{t}^{\text{TLS}_{i}}}
\end{equation}
\noindent where $\beta$ will be gradually increased from $0$ to $1$ during the learning process.
In other words, we let the agent focus on the local tasks first, then use the learned representation to optimize the global behavior.

\subsection{The Deep Neural Network\label{subsec:DeepNeuralNetwork}}

\begin{figure}[htbp]
\begin{centering}
\textsf{\includegraphics[width=1\columnwidth]{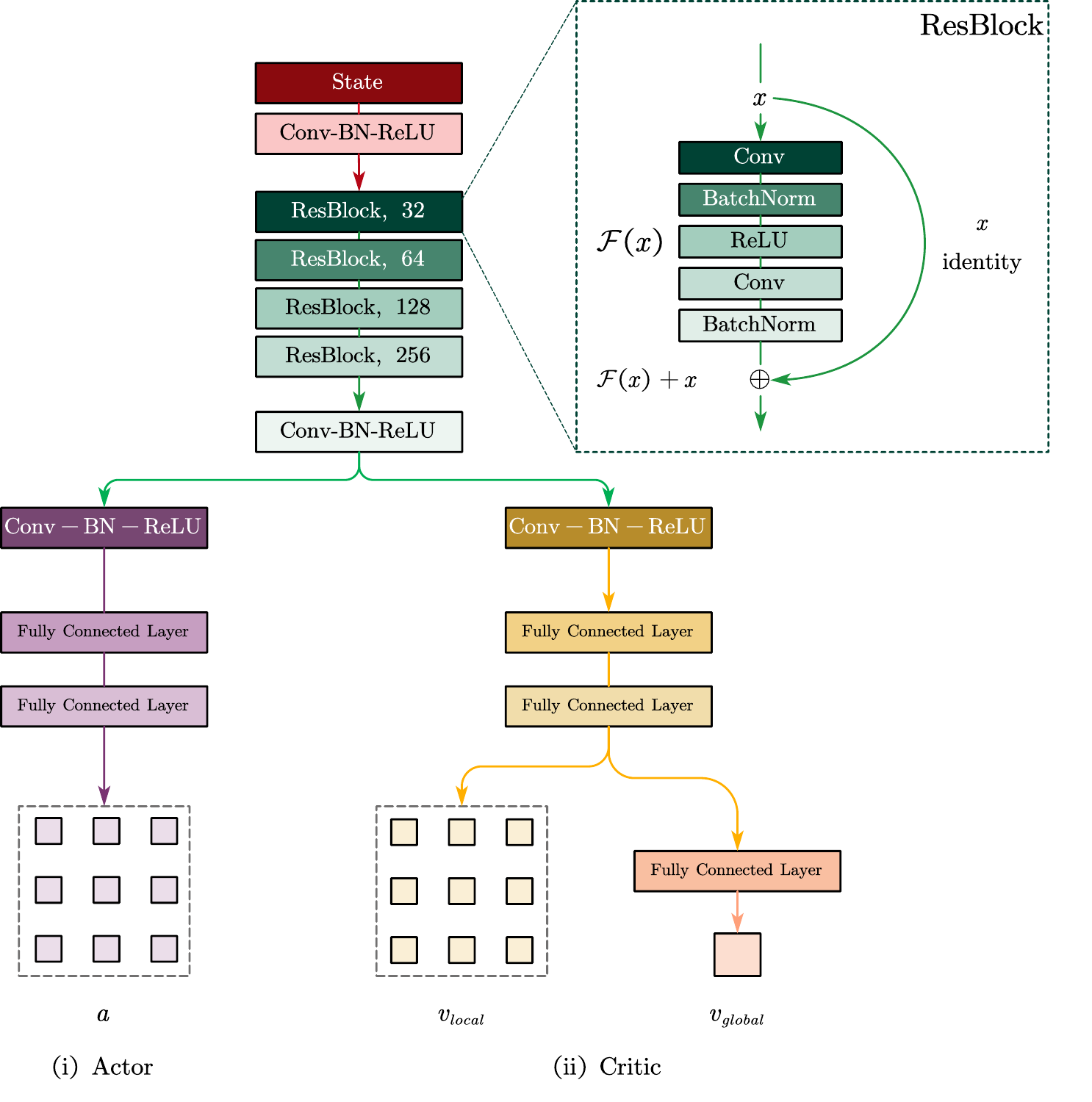}}
\par\end{centering}
\caption{Schematic of the neural network used in this paper.\label{fig:neural-network}}
\end{figure}

In reinforcement learning, we model the agent as an Advantage Actor-Critic (A2C) model. The actor refers to a parameterized policy that defines how actions are selected, and the critic is a method that evaluates each action the agent took. In the context of DRL, both actor and critic are implemented by a deep neural network. The structure of this neural network is demonstrated in Fig.\ref{fig:neural-network}.

The input of the neural network is the state of the system.
The input will be fed into shared layers because it is believed that using shared layers for both actor and critic can bring both knowledge transferring and computational efficiency benefit \cite{Silver-2017}.
In this paper, we use $4$ stacked Residual Blocks \cite{He-2016} as the shared layers,  the output channel of each block is $32$, $64$, $128$, $256$ respectively.

Since we use the A2C model here, we set two separate parts to follow the shared layers. On the left bottom of the figure is the \emph{actor}, which has 2 fully-connected layers. It outputs a $\left\langle N_{\text{TLS}},2\right\rangle $ tensor through a Softmax function, corresponding to the probability to maintain or switch for $N_{\text{TLS}}$ intersections.

On the right bottom of the figure is the \emph{critic}. It contains two fully-connected layers as the actor, but with two separate parts of outputs. There are $N_{\text{TLS}}$ linear scalar outputs in the left side as the \emph{local critic}, which indicates the local value in each intersection.
In the right side, there are two fully-connected layers with one linear scalar output, which is the \emph{global critic} representing the global value.

The outputs of the network are three tensors.
Since there are $9$ intersections in the simulation environment,
the first output is the policy $\pi\left(s_{t}\right)$ in the shape of $\left\langle 9,2\right\rangle $.
The second output is the local critic $v_{\text{local}}\left(s_{t}\right)$ in the shape of $\left\langle 9,1\right\rangle $.
The last output is the global critic $v_{\text{global}}\left(s_{t}\right)$, which is a scalar.

\subsection{Learning Algorithm\label{subsec:Learning Algorithm}}
In general, the parameters of the actor are updated with respect to the critic's evaluation,
and the parameters of the critic are updated with respect to the distance between the evaluation and the real return.
The standard workflow using modern deep learning library is to define two objective functions respectively,
$L^{\text{PG}}$ and $L^{\text{VF}}$  first, then optimize the parameters of networks with respect to them iteratively. The vanilla actor-critic model is hard to train and requires hyper-parameters tuning carefully, due to the data correlation brought by policy-based methods, high sample complexity for critic model optimization,  and the inefficient policy update algorithms. To address these problems, we adopt three recently proposed methods to accelerate the learning speed of the controller.

\underline{First}, we adopt a parallel reinforcement learning paradigm by synchronously training agents on multiple instances of the environment, and update the network averaging over all the actors. Under such a paradigm, the agents will be experiencing a variety of different states and likely to be exploring different parts of the environment at any given time step. Moreover,  we can encourage each actor-learner to use different exploration policies to maximize this diversity. Since the overall changes being made to the parameters by multiple actor-learners applying online updates in parallel are likely to be less correlated in time than a single agent applying online updates, this parallelism can accelerate the exploring speed and decorrelate the data into a stationary process \cite{Mnih-2016}.

In our experiments, each of $N$ (parallel) actors collects $T$ time steps of data in each iteration. Then we construct the objective function on these $NT$ time steps of data and optimize it with Adam \cite{Kingma-2014} algorithm for $K$ epochs.

\underline{Second}, we use an exponentially-weighted estimator of the advantage function, called General Advantage Estimation (GAE) \cite{Schulman-2015a}, to further accelerate the learning process.
As discussed in \emph{Section \ref{sec:rl-background}}, using advantage function can lower variance while estimating the overall sum of return.
However, such an approach typically requires a large number of samples to learn the advantage function. GAE is a recently proposed trick to deal with this problem.
It is  closely analogous to the TD($\lambda$) algorithm\cite{Sutton-1998}. Compared with vanilla advantage estimation algorithm, which will only bootstrap from the (learned) value function for one step (analogous to TD($0$)), GAE can bootstrap for several steps. By increasing the coefficient $\lambda$, such method lower the bias of estimation at the cost of increased variance, and therefore can accelerate the learning speed if $\lambda$ is correctly selected.

Let us define
\begin{equation}\label{eq:delta}
  \delta_{t}^{V}=r_{t}+\gamma V\left(s_{t+1}\right)-V\left(s_{t}\right)
\end{equation}
Since $\delta_{t}^{V}$ is actually an unbiased approximation of advantage at time step $t$, we can therefore consider a series of $k$-step estimate $\hat{A}_{t}^{\left(k\right)}$
\begin{equation}
\left\{\begin{aligned}\hat{A}_{t}^{\left(1\right)} & =\delta_{t}^{V}\\
\hat{A}_{t}^{\left(2\right)} & =\delta_{t}^{V}+\gamma\delta_{t+1}^{V}\\
\dots\\
\hat{A}_{t}^{\left(k\right)} & =\sum_{l=0}^{k-1}\gamma^{l}\delta_{t}^{V}
\end{aligned}\right.
\label{eq:sum_A}
\end{equation}

A truncated version of generalized advantage estimator $\hat{A}_{t}^{GAE\left(\gamma,\lambda\right)}$ can then be defined as the exponentially-weighted average of these $k$-step estimators:
\begin{equation}
\begin{aligned}\hat{A}_{t}^{GAE\left(\gamma,\lambda\right)}: & =\left(1-\lambda\right)\left(\hat{A}_{t}^{\left(1\right)}+\lambda\hat{A}_{t}^{\left(2\right)}+\lambda^{2}\hat{A}_{t}^{\left(3\right)}+\dots\right)\\
 & =\left(1-\lambda\right)\left(\delta_{t}^{V}+\lambda\left(\delta_{t}^{V}+\gamma\delta_{t+1}^{V}\right)+\dot{\dots}\right)\\
 & =\sum_{l=0}^{\infty}\left(\gamma\lambda\right)^{l}\delta_{t+l}^{V}
\end{aligned}
\label{eq:gae_adv}
\end{equation}

We use a simplified notation $\hat{A}_t$ to represent $\hat{A}_{t}^{GAE\left(\gamma,\lambda\right)}$ in following paper.

\underline{Third}, we adopt a new policy update algorithm, called clipped Proximal Policy Optimization (PPO) algorithm \cite{Schulman-2017}.
This algorithm seeks to guarantee a monotonic improvement of stochastic policy by introducing a probability ratios
$\mathbf{r}_{t}\left(\theta\right)=\frac{\pi_{\theta}\left(a_{t}|s_{t}\right)}{\pi_{\theta_{\text{old}}}\left(a_{t}|s_{t}\right)}$,
where $\theta_{\text{old}}$ are the parameters of actor model before current update.

Instead of using advantage directly in policy gradient as mentioned in Eq. \eqref{eq:policy_gradient}, such algorithm uses a  truncated advantage $\text{clip}\left(r_{t}\left(\theta\right),1-\epsilon,1+\epsilon\right)\hat{A_{t}}$. This clip term removes the incentive for moving $r_{t}$ outside of the interval $\left[1-\epsilon,1+\epsilon\right]$, where $\epsilon$ is a hyperparameter that changes during the training process. Such a setting will ignore the change of probability ratio when it would make the objective improve, and only include the change when it makes the objective worse. Then we can construct a surrogate objective function $L^{\text{PG}}$ whose gradient is the policy gradient estimator.

\begin{equation}
L^{\text{PG}}\left(\theta\right)=\hat{\mathbb{E}}\left[r_{t}\left(\theta\right)\hat{A_{t}},\text{clip}\left(r_{t}\left(\theta\right),1-\epsilon,1+\epsilon\right)\hat{A_{t}}\right]\label{eq:ppo_surrogate_loss_pi}
\end{equation}
%
On the other hand, the critic model needs to be trained as well before it can evaluate the value function precisely. The traditional method is to define an objective function $L_{t}^{\text{VF}}$, then optimize the model by the backpropagation algorithm.
Following previous works \cite{Mnih-2016,Schulman-2017}, we define the loss function $L_{t}^{\text{VF}}\left(\theta\right)$ as a squared-error loss between value function and the accumulative return $\left(V_{\theta}\left(s_{t}\right)-R_{t}\right)^{2}$, where $R_{t}$ is calculated according to Eq. \eqref{eq:expected-dis-reward} and \eqref{eq:reward}. For the reason of computational stability, all rewards are normalized into $1$ overall running simulations during the training process.

We modify $L^{\text{VF}}$ into a similar form as $L^{\text{PG}}$ since we use shared layers for both actor and critic. The modified objective function is average over the unclipped squared-error loss and the clip loss.
\begin{equation}\label{eq:ppo_surrogate_loss_v}
\begin{aligned}
  L^{\text{VF}}\left(\theta\right)=&(V_{\tiny{\theta_{\text{old}}}}\left(s_{t}\right)
  +\text{clip}\left[V_{\theta}\left(s_{t}\right)-V_{\tiny{\theta_{\text{old}}}}\left(s_{t}\right),1-\epsilon,1+\epsilon\right]\\
  &-R_{t})^{2}
\end{aligned}
\end{equation}

The objective can further be augmented by adding an entropy bonus to ensure sufficient exploration, as suggested in past works \cite{Williams-1992, Mnih-2016}. The following objective function, which will be \textbf{maximized} at each iteration, can then be obtained by combining all these terms:

\begin{equation}\label{eq:final_obj}
\begin{aligned}
%
&L_{t}\left(\theta\right)=\hat{\mathbb{E}}\left[L_{t}^{\text{PG}}\left(\theta\right)-c_{1}L_{t}^{\text{VF}}\left(\theta\right)+c_{2}S\left[\pi_{\theta}\right]\left(s_{t}\right)\right]\\
\end{aligned}
\end{equation}

Here, $c_{1}$, $c_{2}$ are coefficients of critic loss and entropy bonus, and $S$ denotes the entropy bonus.


Hyperparameters used for following experiments are listed in Table \ref{tab:hyperparameters}, where $\alpha$ is linearly annealed from $1$ to $0$ during the learning process to decay the learning speed.
\begin{table}[htbp]
\caption{Hyperparameters\label{tab:hyperparameters}}
\centering
\begin{tabular}{cc}
\textbf{Hyperparameter} & \textbf{Value}\tabularnewline
\hline
Horizon ($T$)  & $64$\tabularnewline
Learning rate (Adam) & $1.0\times10^{-4}\times\alpha$\tabularnewline
Num. episodes & 50\tabularnewline
Num. epochs  & $3$\tabularnewline
Minibatch size & $64\times16$\tabularnewline
Discount ($\gamma$) & $0.99$\tabularnewline
GAE parameter ($\lambda$)  & $0.95$\tabularnewline
Number of actors & $16$\tabularnewline
Clipping parameter $\epsilon$ & $0.1\times\alpha$\tabularnewline
$L^{\text{VF}}$ coeff. $c_{1}$ in Eq. \eqref{eq:final_obj} & $1.0$\tabularnewline
Entropy coeff. $c_{2}$ in Eq. \eqref{eq:final_obj} & $0.01$\tabularnewline
\end{tabular}
\end{table}

\section{Simulation Results\label{sec:results}}

To validate the effectiveness of the proposed DRL model, we carry out a number of simulation tests.
All experiments were conducted using the traffic micro-simulator SUMO v0.32.0 and its Python API \cite{Krajzewicz-2012}.

\subsection{Traffic Demand Settings for Simulation Tests}

For each instance of simulation, the initial state is a traffic network without any vehicles, then vehicles with a random destination and a corresponding route will be inserted randomly into the network. Each simulation will last for 1 hour (3600 seconds).

Since we seek to propose a method that can be generalized for any situations, the traffic demand is generated randomly via a Binomial distribution $B(b,\frac{1}{np})$ to mimic general cases, where $b$ is the maximum number of simultaneous arrivals and $\frac{1}{p}$ is the expected arrivals in a second. In the training phase, $b$ and $p$ are sampled uniformly from $\left[10, 60\right]$ and $\left[0.1,2\right]$, which means the traffic production is around $1800$ to $36000$ veh/h.

\begin{figure}[h]
\includegraphics[width=1.0\columnwidth]{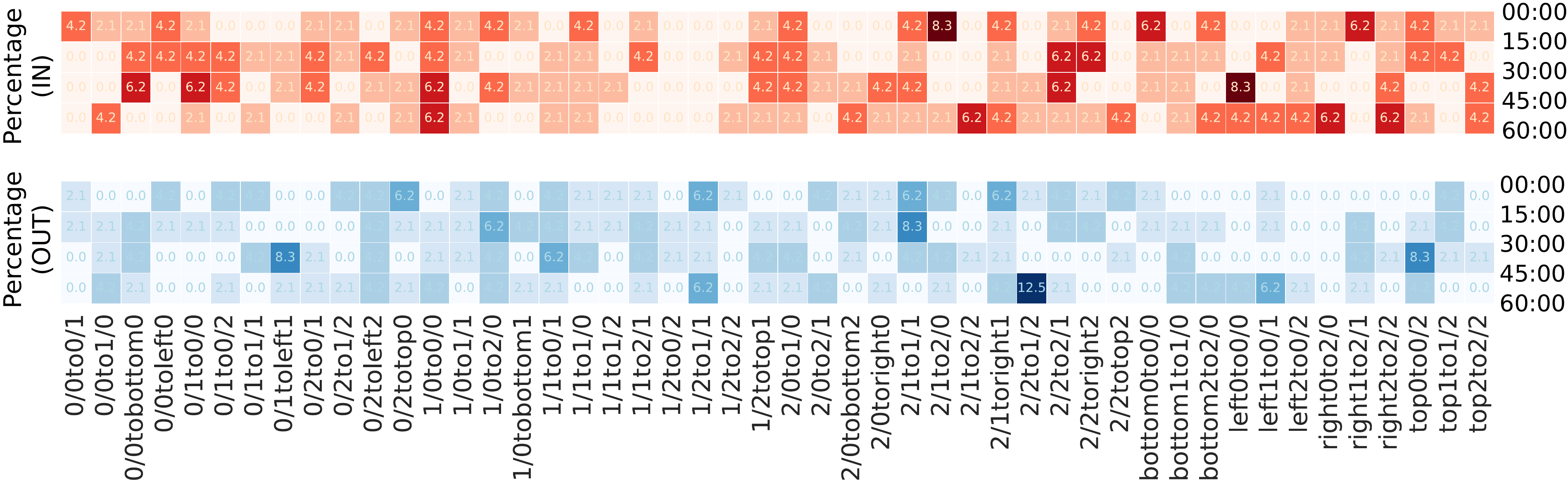}

\caption{An example of randomly generated vehicle routes. The x-axis is the index of each edge, and the y-axis is the periods. The number on each cell indicates the percentage of vehicles entering/leaving the network through the specific edge in that period. For example, in the last period ($45:00 - 60:00$), there is about $25\%$ of total incoming vehicles entering the traffic network from edges \textbf{1/0to1/1, 2/1to2/2, right0to2/0, right2to2/2}, and $12.5\%$ of total incoming
vehicles set the edge \textbf{2/2to1/2} as their destinations. See \cite{-d} for details of implementation.}
\label{fig:random-weights}
\end{figure}

To introduce reasonable randomness, we divide an hour in simulation into $4$ periods. For every $15$ minutes, the routings of vehicles will be alerted. We use two normal distributions to characterize the routings of vehicles. One distribution controls the probabilities that via which edge a vehicle enters the network and the other controls via which edge a vehicle leaves. Such settings can provide directional routes which are often seen in the real traffic scenarios. It is illustrated by an example in Fig. \ref{fig:random-weights}.

\subsection{Performance Comparison}

We compare our method with fixed-time and vehicle-actuated controllers.
In these conventional controllers, the offset of each phase is optimized using Webster Formulation according to the generated trips.
The duration of a phase range from $5$ to $45$ seconds for the vehicle-actuated controller.

The performance is evaluated under three criteria. The first criterion is the number of arrival vehicles, which indicates that for the given period, how many vehicles have arrived at their destination through the controlled area:

\begin{equation}
Arr=\sum_{t=0}^{T}\|\text{Veh}_{\text{out}}\|\label{eq:arrived}
\end{equation}

The second criterion is the average waiting time, indicating the time each vehicle has spent in halting speed in average:

\begin{equation}
\overline{T}_{wait}=\hat{\mathbb{E}}\left[\sum_{i=0}^{N}T_{wait}^{\left(\text{veh}_{i}\right)}\right]\label{eq:waiting}
\end{equation}

The third criterion is the time loss, which is the gap between the ideal time and actual time it spends to arrive at its destination:

\begin{equation}
\overline{T}_{loss}=\hat{\mathbb{E}}\left[\sum_{i=0}^{N}\left(T_{real}^{\left(\text{veh}_{i}\right)}-T_{ideal}^{\left(\text{veh}_{i}\right)}\right)\right]\label{eq:timeloss}
\end{equation}

For the criterion $\overline{T}_{wait}$ and $\overline{T}_{loss}$, we only consider vehicles that have arrived at its destination.

\subsection{Training Speed}

In this paper, the agent is built with PyTorch \cite{Paszke-2017} and communicates with simulation environment via the Traci library \cite{Wegener-2008}. Both simulations and deep learning process are run on a workstation with Intel Core i7-6700K CPU, 32GB RAM and 2 Nvidia GeForce Titan X GPUs.


As mentioned in Table \ref{tab:hyperparameters}, the agent is trained for 50 episodes, and each episode has 3600 simulation steps.
In each episode, the DRL model will be updated for $3$ epochs at every 128 simulation steps.
An epoch is a single pass through the entire training set, followed by testing of the verification set.
That means the whole training process contains 180000 forward passes ($50\times3600$, for traffic lights control) and 4219 backward passes ($50\times3600/128\times3 $, for neural network update).
The total process lasts about 7 hours 30 minutes on our workstation.

\begin{figure}
\begin{centering}
\includegraphics[width=0.9\columnwidth]{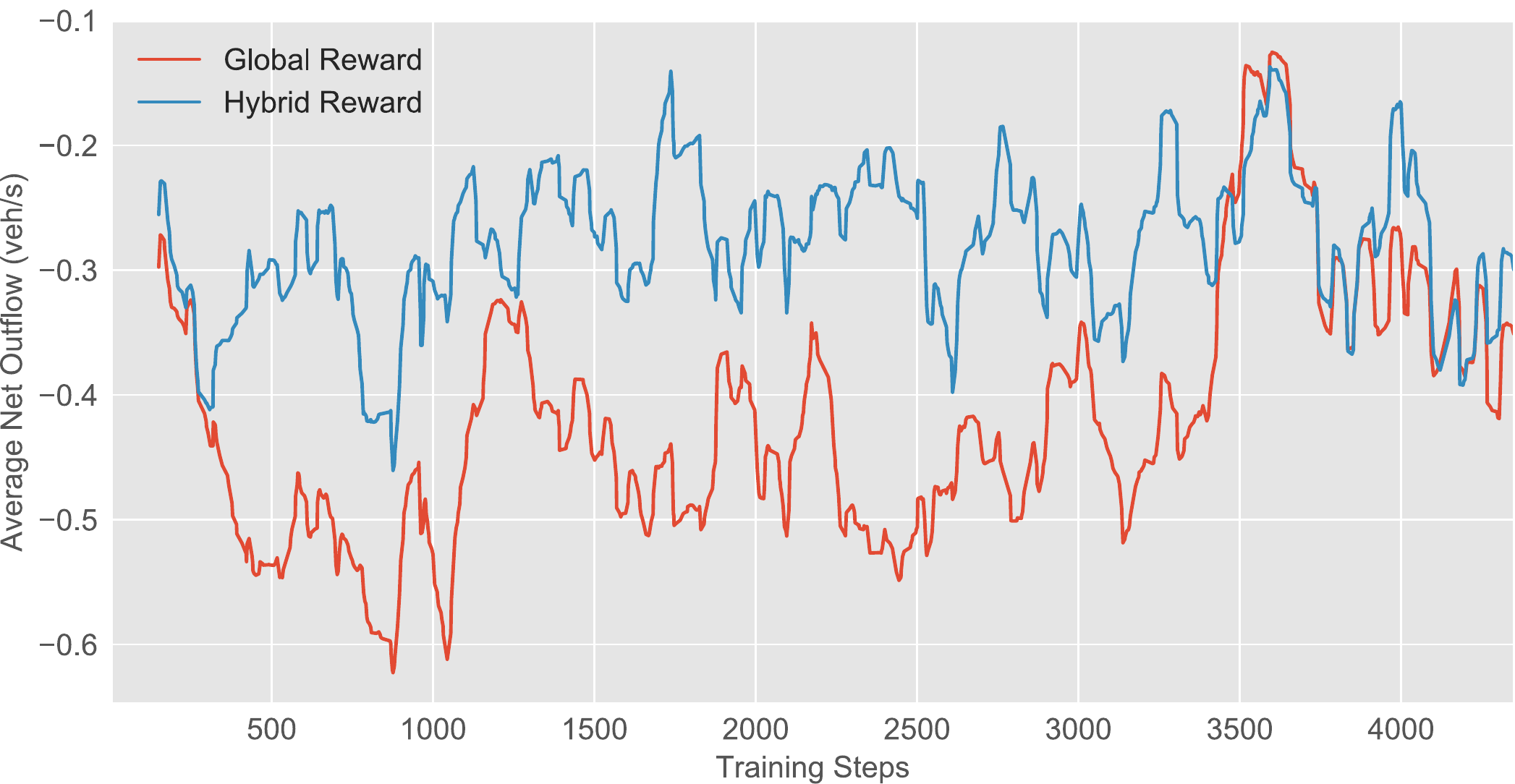}
\end{centering}
\caption{Average net outflow during training process. \label{fig:training-procedure}}
\end{figure}

As illustrated in Fig. \ref{fig:training-procedure}, we test two models with the same structure,
except one updated by only global reward (the net outflow),
while another by the hybrid reward that includes both global reward and local reward (the opposite of absolute imbalance of each intersection).
We can see that the one using hybrid reward achieves significantly better performance than the other one within a few episodes.

\subsection{Performance Comparison}

We compare different controllers on 360 different traffic demand settings.
There are $6$ different kinds of traffic demands ranging from $1800$ to $36000$ veh/h and $6$ different randomness, i.e., $b=10, 20, 30, 40, 50, 60$.
Such setting forms 36 combinations with different traffic demand and randomness.
For each combination, there are $10$ simulations generated.

\begin{figure*}
\begin{centering}
\subfloat[Arrived vehicles]{\includegraphics[width=0.32\textwidth]{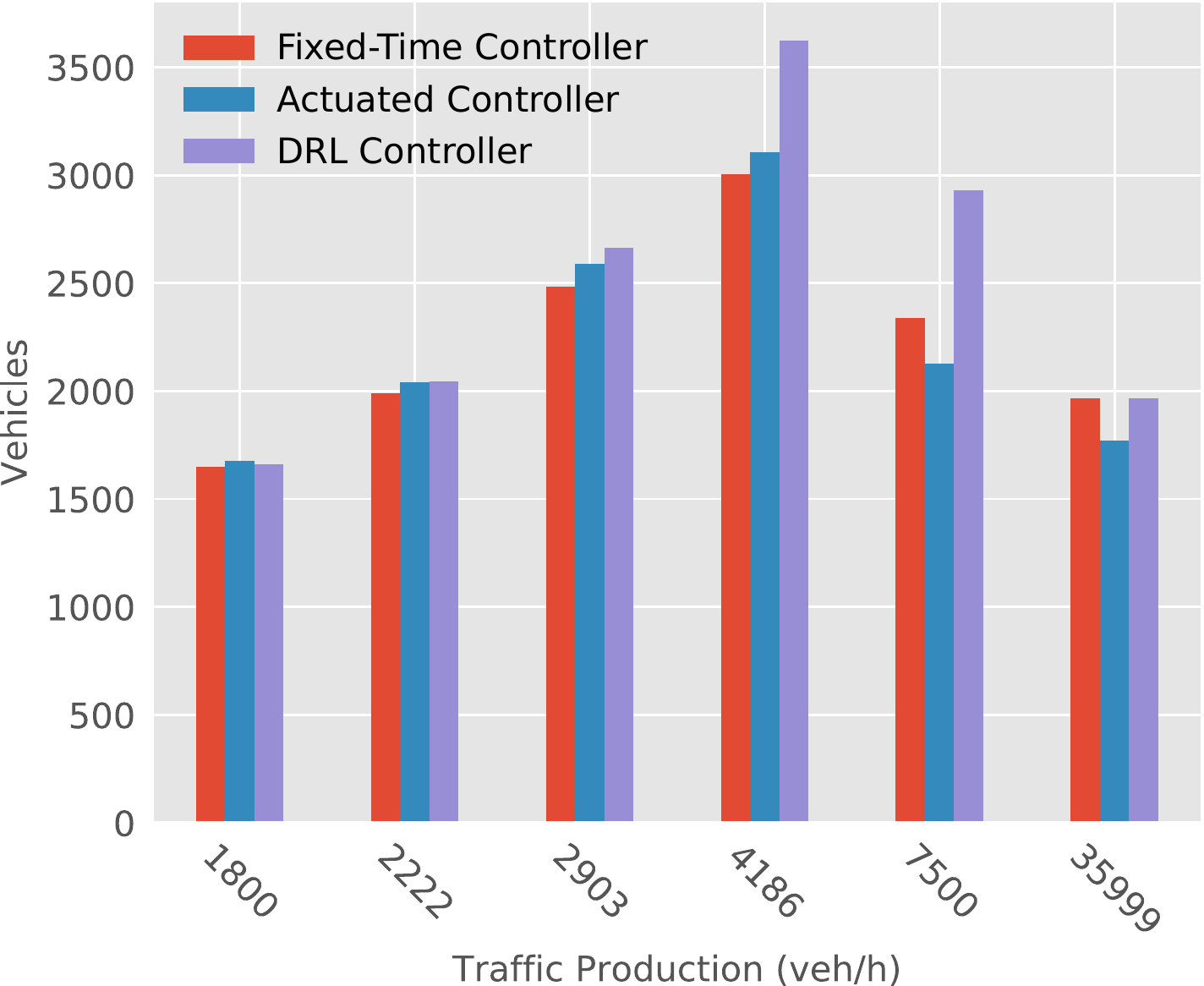}}
\hspace{0.005\textwidth}
\subfloat[Waiting time]{\includegraphics[width=0.32\textwidth]{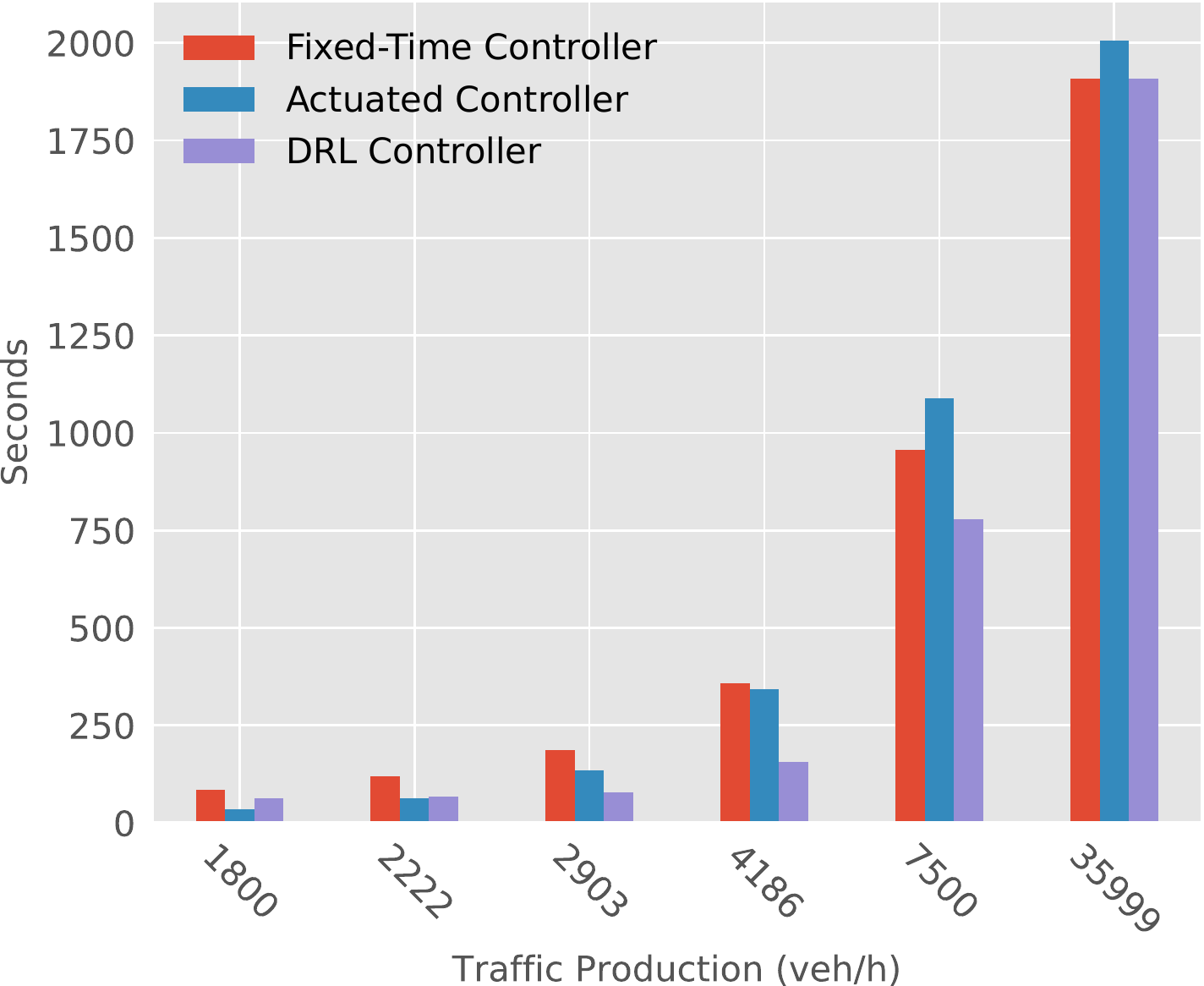}}
\hspace{0.005\textwidth}
\subfloat[Time losses]{\includegraphics[width=0.32\textwidth]{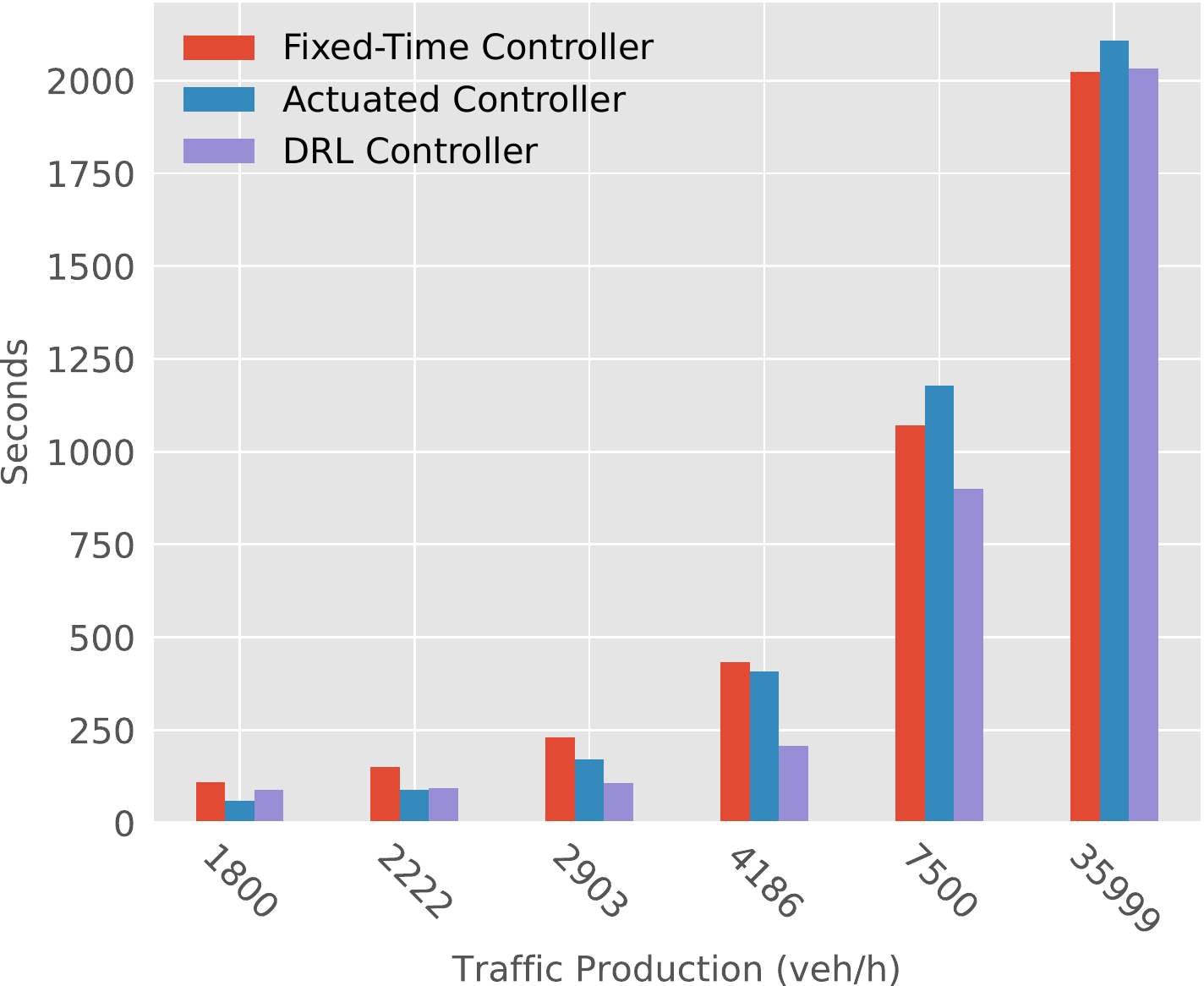}}
\end{centering}
\caption{Average performance of different controllers. \label{fig:Performance-of-DUTCS}}
\end{figure*}

Fig. \ref{fig:Performance-of-DUTCS} shows the average performance of different controllers under the given traffic demands. As the experiments have shown, DRL based method is advanced than fixed-time and vehicle-actuated controllers in unsaturated and saturated cases, but its performance is getting close to the fixed-time controller once the traffic system becomes over-saturated. Among all situations, the average throughput of traffic system increases by $25.19\%$ and $37.81\%$ at maximum compared with fixed-time and vehicle-actuated controllers, while the average waiting time reduces by $18.68\%$ and $28.54\%$ at the same time. More detail results can be found in Appendix \ref{sec:performance-tables}.

\begin{figure*}
\begin{centering}
\subfloat[Unsaturated situation($2400$ veh/h)]{\includegraphics[width=0.32\textwidth]{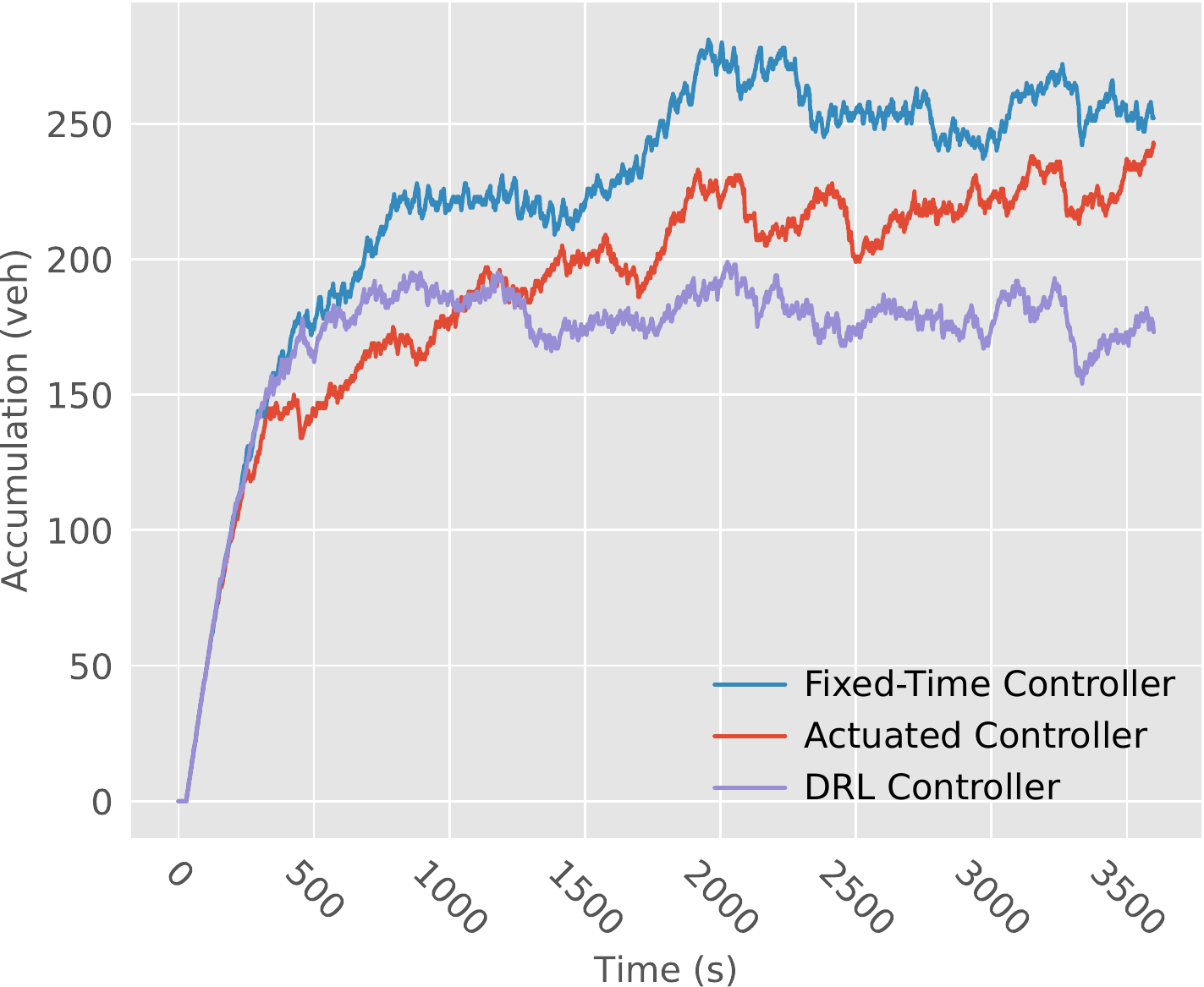}}
\hspace{0.005\textwidth}
\subfloat[Saturated situation($3600$ veh/h)]{\includegraphics[width=0.32\textwidth]{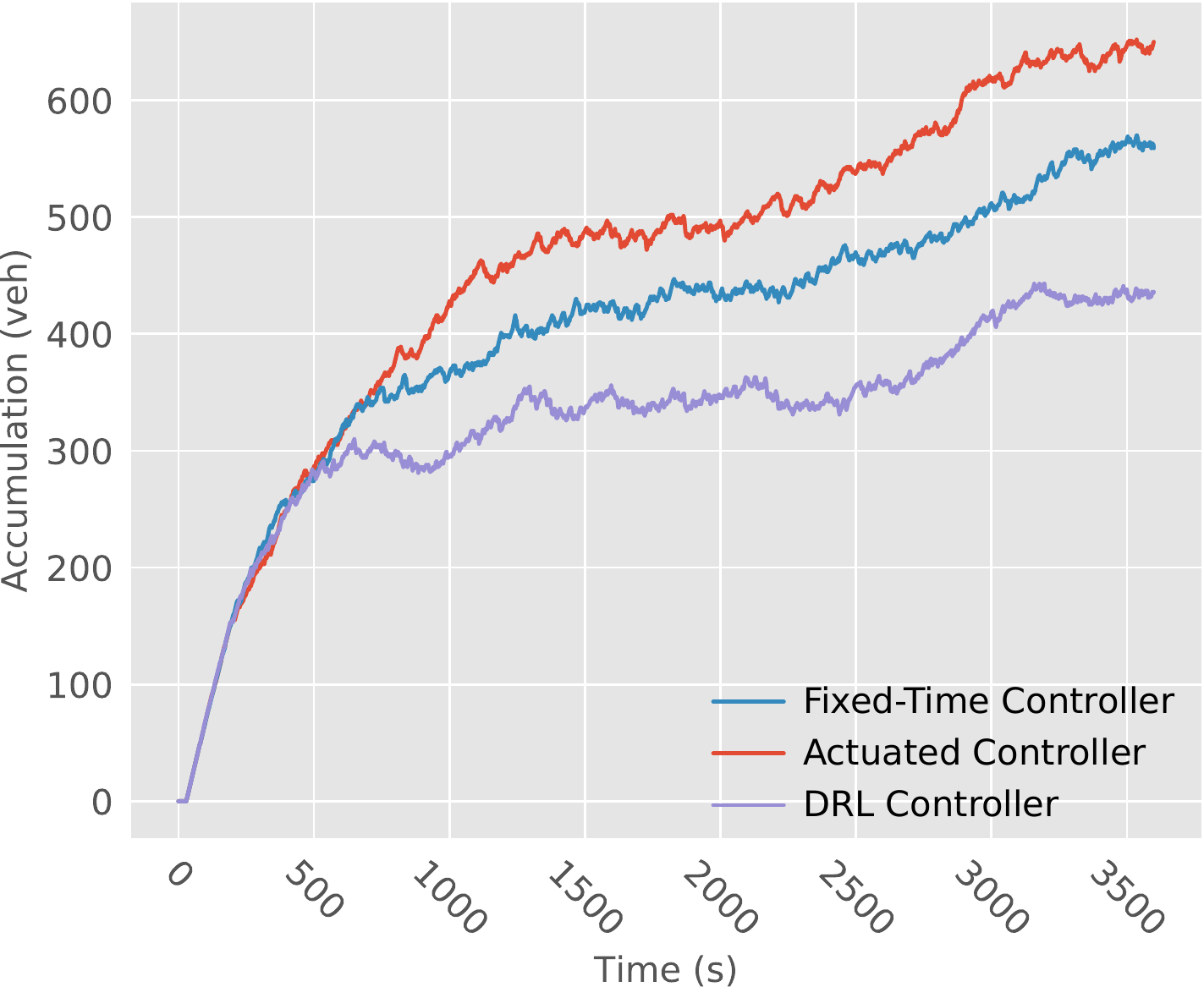}}
\hspace{0.005\textwidth}
\subfloat[Over-saturated situation($7200$ veh/h)]{\includegraphics[width=0.32\textwidth]{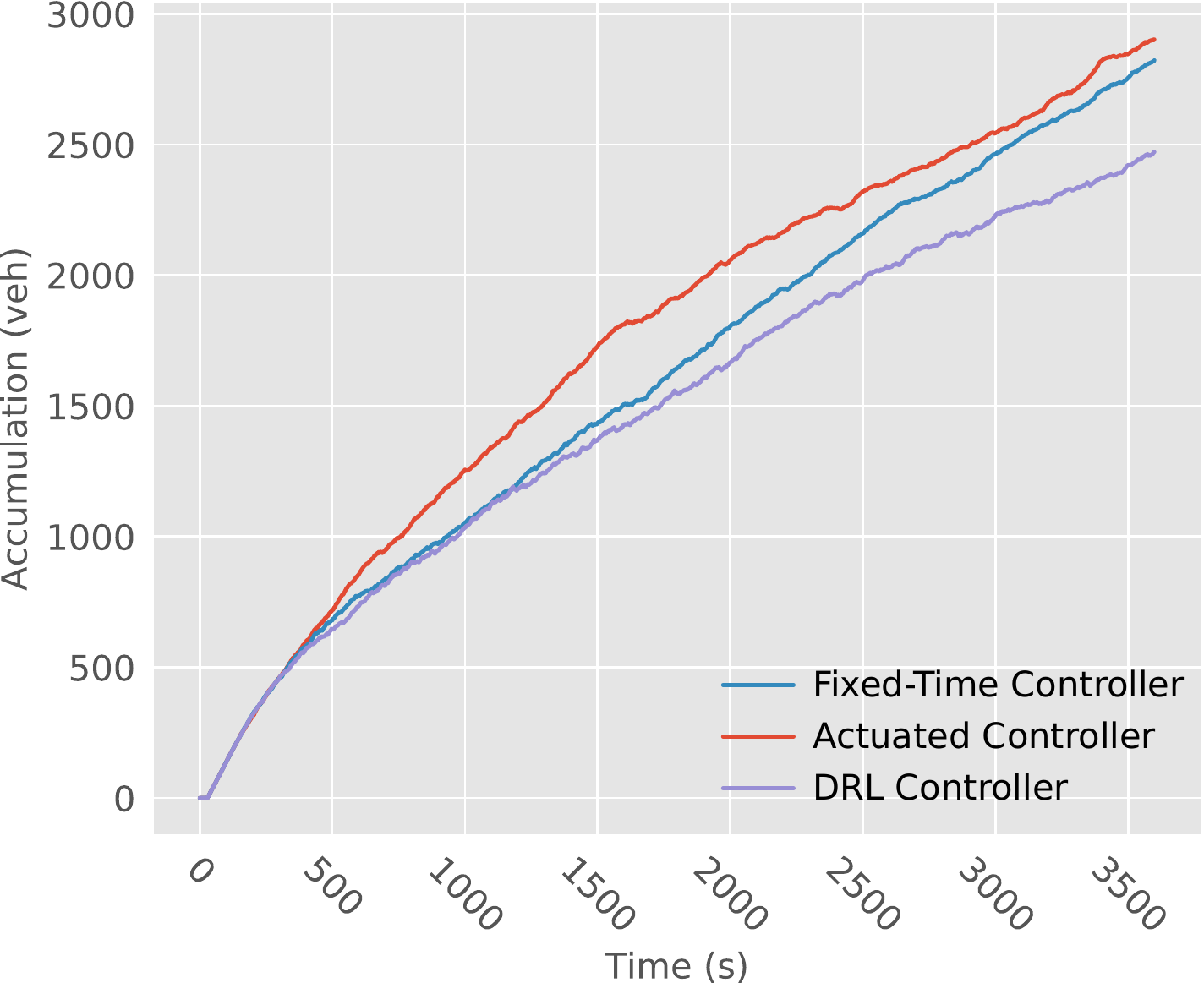}}
\end{centering}
\caption{Macroscopic fundamental diagrams for different controllers. The x-axis is the simulation steps, and the y-axis indicates the number of vehicles in the traffic grid. \label{fig:mfd}}
\end{figure*}

To better understand the experiment results, we draw the Macroscopic Fundamental Diagrams (MFD) \cite{Geroliminis-2007, Daganzo-2008,Geroliminis-2008} for three typical traffic demand settings when different controllers are applied. Fig. \ref{fig:mfd} have shown that, for all these traffic conditions, the traffic accumulation (the number of vehicles in the traffic network) is the lowest under the control of DRL strategies, so is its increasing rate.

Such phenomenon prove our DRL controller could better evacuate the vehicles-in-net than traditional controllers. Unlike the vehicle-actuated controller which performs well only in the unsaturated situations, the DRL controller outperforms the fixed-time controller in all situations. It is believed that such phenomenon happened because the vehicle-actuated controller can only be implemented to an isolated intersection\cite{PengJing-2017}. Due to such short-sightseeing, vehicle-actuated controller leads the traffic system to a local optimum. In contrast, our DRL controller considers the global state to make better decisions and thus achieves better performance.

\section{Conclusions\label{sec:conclusions}}

In this paper, we propose an efficient DRL based approach for UTC.
The simulation experiments have shown that our method performs better than tradition UTC approaches and can handle more complex environments while using fewer computing resources.

It should be pointed out that there are several things to be fathomed for this new DRL model. For example, how to transform the state into a proper format for the more general, unstructured traffic network might be one of the most urgent problems needed to be discussed. In addition, whether we should apply some other neural networks for better performance needs to be answered. We hope that this paper can provide a good start point for the following studies and expect to obtain new achievements in the near future.

\bibliographystyle{IEEEtran}
\bibliography{IEEEabrv,citations}

\newpage
\appendices
%


\section{Average performance in Different Situations\label{sec:performance-tables}}

\begin{table}[htbp]
  \centering
  \caption{Arrived vehicles for different traffic production}
    \begin{tabular}{l|lll}
    \multicolumn{1}{p{0.125\columnwidth}}{Traffic Demand} & \multicolumn{1}{p{0.15\columnwidth}}{Fixed-Time Controller} & \multicolumn{1}{p{0.15\columnwidth}}{Actuated Controller} & \multicolumn{1}{p{0.15\columnwidth}}{RL Controller} \\
    \hline
    1800  & 1648.65  & \textbf{1677.93 } & 1661.65  \\
    2222  & 1988.47  & 2042.02  & \textbf{2045.85 } \\
    2903  & 2482.17  & 2589.30  & \textbf{2664.23 } \\
    4186  & 3004.28  & 3107.12  & \textbf{3623.52 } \\
    7500  & 2339.50  & 2125.35  & \textbf{2928.90 } \\
    36000  & 1964.62  & 1771.48  & \textbf{1968.42 } \\
    \hline
    Average &2237.95 	&2218.87 	&\textbf{2482.09}\\
    \end{tabular}%
  \label{tab:addlabel}%
\end{table}%

\begin{table}[htbp]
  \centering
  \caption{Waiting time for different traffic production}
    \begin{tabular}{l|lll}
    \multicolumn{1}{p{0.125\columnwidth}}{Traffic Demand} & \multicolumn{1}{p{0.15\columnwidth}}{Fixed-Time Controller} & \multicolumn{1}{p{0.15\columnwidth}}{Actuated Controller} & \multicolumn{1}{p{0.15\columnwidth}}{RL Controller} \\
    \hline
    1800  & 83.86  & \textbf{35.22 } & 63.62  \\
    2222  & 118.97  & \textbf{63.71 } & 67.54  \\
    2903  & 187.12  & 135.40  & \textbf{77.76 } \\
    4186  & 357.60  & 343.59  & \textbf{155.24 } \\
    7500  & 957.42  & 1089.52  & \textbf{778.53 } \\
    36000  & 1909.33  & 2006.02  & \textbf{1908.98 } \\
    \hline
    Average & 602.38  & 612.24  & \textbf{508.61 } \\
    \end{tabular}%
  \label{tab:addlabel}%
\end{table}%

\begin{table}[htbp]
  \centering
  \caption{Time loss for different traffic production}
    \begin{tabular}{l|lll}
    \multicolumn{1}{p{0.125\columnwidth}}{Traffic Demand} & \multicolumn{1}{p{0.15\columnwidth}}{Fixed-Time Controller} & \multicolumn{1}{p{0.15\columnwidth}}{Actuated Controller} & \multicolumn{1}{p{0.15\columnwidth}}{RL Controller} \\
    \hline
    1800  & 110.54  & \textbf{58.40 } & 88.96  \\
    2222  & 150.62  & \textbf{89.98 } & 94.37  \\
    2903  & 229.47  & 170.17  & \textbf{108.04 } \\
    4186  & 432.99  & 408.04  & \textbf{208.11 } \\
    7500  & 1071.85  & 1178.03  & \textbf{901.13 } \\
    36000  & \textbf{2024.27 } & 2107.58  & 2033.29  \\
    \hline
    Average & 669.96  & 668.70  & \textbf{572.32 } \\
    \end{tabular}%
  \label{tab:addlabel}%
\end{table}%

\begin{table}[htbp]
  \centering
  \caption{Arrived vehicles for different randomness}
    \begin{tabular}{l|lll}
    \multicolumn{1}{p{0.15\columnwidth}}{Randomness $b$} & \multicolumn{1}{p{0.15\columnwidth}}{Fixed-Time Controller} & \multicolumn{1}{p{0.15\columnwidth}}{Actuated Controller} & \multicolumn{1}{p{0.15\columnwidth}}{RL Controller} \\
    \hline
    10    & 2284.63  & 2247.12  & \textbf{2535.43 } \\
    20    & 2237.17  & 2212.35  & \textbf{2477.45 } \\
    30    & 2206.18  & 2197.73  & \textbf{2468.25 } \\
    40    & 2234.57  & 2226.12  & \textbf{2460.35 } \\
    50    & 2258.28  & 2237.42  & \textbf{2499.20 } \\
    60    & 2206.85  & 2192.47  & \textbf{2451.88 } \\
    \hline
    Average & 2237.95  & 2218.87  & \textbf{2482.09 } \\
    \end{tabular}%
  \label{tab:addlabel}%
\end{table}%

\begin{table}[htbp]
  \centering
  \caption{Waiting time for different randomness}
    \begin{tabular}{l|lll}
    \multicolumn{1}{p{0.15\columnwidth}}{Randomness $b$} & \multicolumn{1}{p{0.15\columnwidth}}{Fixed-Time Controller} & \multicolumn{1}{p{0.15\columnwidth}}{Actuated Controller} & \multicolumn{1}{p{0.15\columnwidth}}{RL Controller} \\
    \hline
    10    & 590.86  & 608.23  & \textbf{499.11 } \\
    20    & 606.57  & 621.20  & \textbf{515.40 } \\
    30    & 612.76  & 618.29  & \textbf{512.77 } \\
    40    & 610.04  & 617.10  & \textbf{519.57 } \\
    50    & 589.51  & 602.18  & \textbf{498.75 } \\
    60    & 604.54  & 606.45  & \textbf{506.07 } \\
    \hline
    Average & 602.38  & 612.24  & \textbf{508.61 } \\
    \end{tabular}%
  \label{tab:addlabel}%
\end{table}%

\begin{table}[htbp]
  \centering
  \caption{Time loss for different randomness}
    \begin{tabular}{l|lll}
    \multicolumn{1}{p{0.15\columnwidth}}{\textbf{Randomness $b$}} & \multicolumn{1}{p{0.15\columnwidth}}{Fixed-Time Controller} & \multicolumn{1}{p{0.15\columnwidth}}{Actuated Controller} & \multicolumn{1}{p{0.15\columnwidth}}{RL Controller} \\
    \hline
    10    & 659.34  & 665.86  & \textbf{564.34 } \\
    20    & 674.15  & 676.97  & \textbf{577.78 } \\
    30    & 681.18  & 675.09  & \textbf{576.14 } \\
    40    & 674.72  & 672.30  & \textbf{580.53 } \\
    50    & 657.54  & 659.99  & \textbf{565.13 } \\
    60    & 672.80  & 661.99  & \textbf{569.98 } \\
    \hline
    Average & 669.96  & 668.70  & \textbf{572.32 } \\
    \end{tabular}%
  \label{tab:addlabel}%
\end{table}%

\end{document}